\documentclass[journal]{IEEEtran}
\usepackage{amssymb}
\usepackage[algoruled,vlined,linesnumbered]{algorithm2e}
\usepackage{comment}
\usepackage{amsthm}

\theoremstyle{definition}

\theoremstyle{remark}

\usepackage{bm}
\usepackage{cite} 
\usepackage{amsmath}
\usepackage{float}
\usepackage{graphicx}
\usepackage{subfigure}
\usepackage{xcolor}

\begin{document}
\title{Coupled Network for Robust Pedestrian Detection with Gated Multi-Layer Feature Extraction and Deformable Occlusion Handling}

\author{Tianrui Liu,
        Wenhan Luo,
        Lin Ma,
        Jun-Jie Huang,
        Tania Stathaki, and~Tianhong Dai 
        }

\maketitle

\begin{abstract}
Pedestrian detection methods have been significantly improved with the development of deep convolutional neural networks. Nevertheless, detecting small-scaled pedestrians and occluded pedestrians remains a challenging problem. In this paper, we propose a pedestrian detection method with a couple-network to simultaneously address these two issues. One of the sub-networks, the gated multi-layer feature extraction sub-network, aims to adaptively generate discriminative features for pedestrian candidates in order to robustly detect pedestrians with large variations on scale. The second sub-network targets in handling the occlusion problem of pedestrian detection by using deformable regional RoI-pooling. We investigate two different gate units for the gated sub-network, namely, the channel-wise gate unit and the spatio-wise gate unit, which can enhance the representation ability of the regional convolutional features among the channel dimensions or across the spatial domain, repetitively. Ablation studies have validated the effectiveness of both the proposed gated multi-layer feature extraction sub-network and the deformable occlusion handling sub-network. With the coupled framework, our proposed pedestrian detector achieves the state-of-the-art results on the Caltech and the CityPersons pedestrian detection benchmarks.

\end{abstract}


\IEEEpeerreviewmaketitle

\section{Introduction}
\label{sec:introduction}

\IEEEPARstart{P}{edestrian} detection has long been an attractive topic in computer vision and has significant impact on both research and industry. Pedestrian detection is essential to the understanding of a scene, and has a wide range of applications such as video surveillance, robotics automation and intelligent driving assistance systems, while it remains a challenging task.
Pedestrians are usually within a complex environment, like for example images captured by an intelligence vehicle driving in urban scenarios mostly having changing and complex backgrounds. 
Apart from the complex backgrounds, the pedestrian detection task is also challenged by a large variety of poses and appearances in real life scenarios. 
Depending on how close the pedestrian is to the camera, the size of pedestrians in the captured images can vary within a large range.
Partial occlusions between pedestrians and their surrounding instances under crowded environments further raise additional challenges to this task.



Recent advances of Deep Neural Networks (DNNs) have made significant improvements on the performance of the pedestrian detection task \cite{IsfasterRCNNPed,CityPersons, 2018scale-awareRCNN, Tianrui-BMVC2018, Occ-awareRCNN2018}. The detection difficulties due to complex environment and high variances on appearances can be largely resolved by powerful feature representations generated by DNNs.
Nevertheless, it still remains challenging to detect pedestrians of small size or being heavily occluded. For small pedestrians, the image resolution is relatively low, therefore there is less visual information which can be unitized for feature representation and subsequent classification and location estimation. For detecting pedestrians with occlusions, if a rigid detection model is used for the holistic body structure, normally a very low confidence score will be obtained for the detecting window which may lead to missing detections.

Fusing global and local information together has been proven to be effective in many visual tasks. The Deformable Part Models (DPM) detector \cite{DPM} is a successful example of incorporating a global model and local part models. 
Similarly, Zhao \textit{et al.} \cite{pyramidSceneParsing2017} designed a pyramid pooling module to effectively extract a hierarchical global contextual prior, and then concatenated it with the local convolutional network features to improve the performance on scene parsing tasks. In \cite{couplenet}, two sibling networks are coupled together in which one sibling network targets on extracting global information and the other one targets on extracting local information. The global sibling network applies region of interest (RoI) pooling \cite{fasterRCNN2015} to predict a global score of this RoI, and  the local sibling network uses position-sensitive region of interest (PSRoI) pooling \cite{RFCN} to generate score maps which are sensitive to local parts of the object. By coupling the global confidence with the local part confidence together, one can obtain a more reliable prediction.

Targeting on developing a robust pedestrian detection method which can simultaneously address the problems of detecting pedestrians of small size or with partial occlusions, we propose a pedestrian detection framework with two coupled networks 
where the gated multi-layer feature extraction sub-network aims to adaptively generate discriminative features for different pedestrian candidates in order to robustly detect pedestrians with large variations on scales, and the deformable occlusion handling sub-network targets on handling the occlusion problem of pedestrian detection by using deformable regional RoI-pooling.

The proposed gated multi-layer feature extraction sub-network consists of squeeze networks and gate networks.
A squeeze network is applied to reduce the dimension of RoI feature maps pooled from each convolutional layer. It is an essential component in the gated multi-layer feature extraction network which helpd to achieve a good balance between performance and model complexity. What follows is a gate network applied to the feature maps to decide whether features from this layer are essential for representing this RoI. 
We investigate two gate units to manipulate and select features from multiple layers of a DNN, namely, a spatio-wise gate unit and a channel-wise gate unit. 
The expectation is that features manipulated by the proposed two gate units should be able to possess stronger inter-dependencies among channels and among spatial locations, respectively.


In the deformable occlusion handling sub-network, we propose to use deformable regional RoI-pooling which can better fit the non-rigid parts of human body than using the traditional RoI-pooling \cite{fasterRCNN2015}. The deformable occlusion handling network is fully convolutional and with deformable RoI pooling. Deformable RoI-pooling uses shiftable pooling grids which can adapt to the local parts of deformable pedestrians. The offsets of the shifting pooling grids are learnt through additional convolutional layers in the deformable occlusion handling sub-network. 

The contributions of this work are as follows.
First, we propose a coupled framework for pedestrian detection which can address problem of detecting small pedestrians and detecting pedestrians with occlusions simultaneously. 
Second, the proposed gated feature extraction sub-network can adaptively extract multi-layer convolutional features for pedestrian candidates. The two different types of gate units we proposed can manipulate the RoI feature maps in the channel-wise and spatio-wise manner, respectively.
Thirdly, the deformable occlusion handling sub-network enables more robust detection through a deformable RoI-pooling which can adaptively adjust the relative positions of the pooling grids for body parts which can better fit the deformable/occluded pedestrians.
With the coupled detection framework, the two sub-networks use two complimentary ways of detection to reinforce the robust detection results, leading to state-of-the-art performance on the challenging CityPersons and Caltech pedestrian datasets.


 
\section{Related Works}
\label{sec:relatedworks}

\subsection{Traditional Pedestrian Detection Methods}

Traditional features that have been exploited for pedestrian detection include Haar-like features \cite{Haar-ped1997},  Scale-Invariant Feature Transform (SIFT) \cite{SIFT}, Local Binary Pattern (LBP) \cite{LBP-2002PAMI}, edgelets \cite{DetectionandTracking-2007IJCV}, and Histogram of Oriented Gradient (HOG) \cite{HOG}, etc. Among them, HOG and its variations, such as ACF \cite{ACF_2014dollar2014}, LDCF \cite{FilterChannal-2015} and Checkerboards \cite{HowFar-2016}, are considered as arguably the most successful hand-engineered features for pedestrian detection. These features are used in conjunction with a classifier, for instance boosted forests, to perform pedestrian detection via classification. 

Traditional detection approaches \cite{DPM,2Ped-PAMI,2PedTrack-IJCV-SiyuTang} which use a holistic model for the entire pedestrian body structure normally do not work well when the pedestrian is partially occluded. Deformable Part Models (DPM) \cite{DPM} is one of the most successful early attempts on occlusion handling for pedestrian detection. It applies a part-based model which contains a root-filter (analogous to the HOG filter) and a set of part-filters associated with deformation costs measuring the deviation of each part from its ideal location. Ouyang \textit{et al.} \cite{2Ped-PAMI} extended the DPM framework for crowded pedestrian detection by using a two-pedestrian detector to reinforce the detection scores of the single-pedestrian detector. Similarly, a set of occlusion patterns of pedestrians have been explored in \cite{2PedTrack-IJCV-SiyuTang} to deal with inter-pedestrian occlusions.

The above methods were the dominant approaches for pedestrian detection before the emerging of deep neural networks based pedestrian detection methods.

\subsection{Deep Neural Network based Pedestrian Detection Methods}

Deep Neural Networks (DNNs) based pedestrian detection methods \cite{IsfasterRCNNPed,CityPersons, 2018scale-awareRCNN, MCF_TIP2017,  Small_ped_TTL2018, fasterRCNN2015, PDOE_BiboxRegression_eccv2018,Tianrui-BMVC2018, repulsion_loss2018, occluded_shanshan2018,shanshan_occ_ped_cvpr2018,GDFL_eccv2018} have demonstrated superior detection performance. One of the main reasons is that DNNs are able to generate features with a stronger discrimination capability through end-to-end training when compared to the hand-engineered feature representations. 

Faster Region-based Convolutional Neural Networks (Faster-RCNN) \cite{fasterRCNN2015} have achieved an excellent detection accuracy in general object detection tasks. Many recent detection methods \cite{IsfasterRCNNPed,CityPersons, 2018scale-awareRCNN, Tianrui-BMVC2018} are variations of them. 
RCNN-based methods \cite{RegionCNN-2014CVPR,fastrcnn15,fasterRCNN2015} make use of a two-stage detection strategy. That is, a small number of highly potential candidate regions are first proposed using the Region Proposal Network (RPN) and then classification is performed based on the extracted CNN features from these candidate regions. Directly applying Faster-RCNN for pedestrian detection \cite{IsfasterRCNNPed,CityPersons} does not lead to satisfying results since small pedestrians which dominate common pedestrian datasets \cite{PedBenchmark-2009CVPR-Dollar,CityPersons} are usually not well detected. Zhang \textit{et al.} \cite{CityPersons} tailored Faster-RCNN in terms of detection anchors to accommodate for the pedestrian detection task. However, the low feature resolution at a deep layer limits pedestrian detectors from accurately detecting pedestrians with small size. To take advantage of the CNN features from multiple layers, the Multi-layer Channel Features (MCF) \cite{MCF_TIP2017} method and the Region Proposal Network with Boosted Forests (RPN+BF) \cite{IsfasterRCNNPed} method propose to concatenate features from multiple layers of a CNN and replace the downstream classifier of Faster-RCNN with boosted forests to improve performance on detecting hard samples, while a fixed feature combination is used for all pedestrians.


\begin{figure*}[t]
\centering \includegraphics[width=1.9\columnwidth]{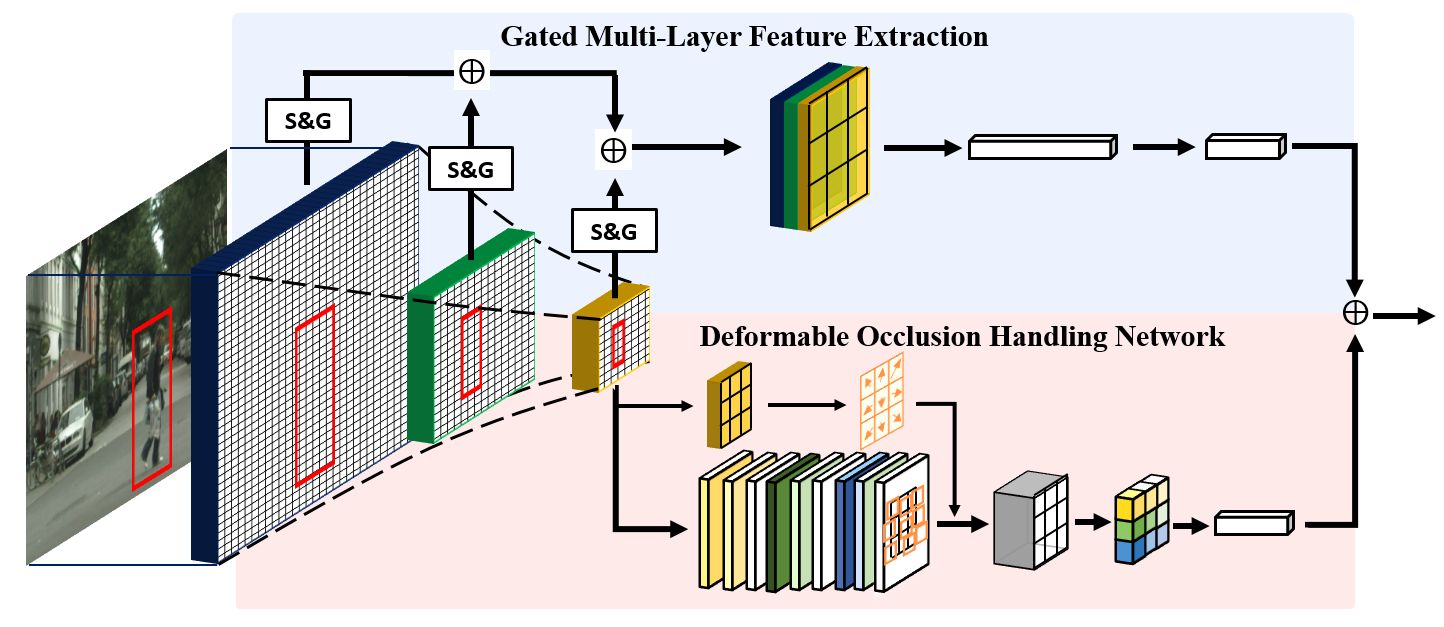}
\caption{Overview of the proposed pedestrian detection framework. The backbone network is VGG16 \cite{VGG16} which has 5 convolutional blocks. For illustration purposes, only 3 convolutional blocks are shown in this figure.
The proposed pedestrian detection framework contains two coupled sub-networks: the upper sub-network is a gated multi-layer feature extraction network which can provide highly discriminative CNN features especially for small pedestrians under complex backgrounds; the lower sub-network is a deformable occlusion handling stream which applies deformable RoI-pooling and is used to generate robust features for occluded pedestrians. }
\label{fig:coupled_network} 
\end{figure*}

There are also DNN-based methods in the literature targeting on the problem of detecting pedestrians with occlusions. DeepPyramid DPM method \cite{DPMareCNN} utilizes features extracted from a learned CNN feature extractor and has shown improved performance compared to the traditional DPM-based methods \cite{DPM,2Ped-PAMI,2PedTrack-IJCV-SiyuTang}. A multi-label learning method \cite{Multilabel_OCC_ped_ICCV2017} is proposed to improve the performance of part detectors and reduce the computational cost of integrating multiple part scores. However, the body parts used in these methods are manually designed, which may limit their performance. Some recent pedestrian detection methods \cite{repulsion_loss2018,Occ-awareRCNN2018} introduce novel loss functions to solve the occlusion problem when detecting pedestrians in crowded scenarios. In occlusion-aware RCNN \cite{Occ-awareRCNN2018},  an occlusion-aware region-of-interest (RoI) pooling unit is proposed to integrate the visible prediction into the network for occlusion handling. Five part anchors at hand-selected positions are used for pooling the regional features of five body parts. The problem is that pedestrians are not rigid object and can deform under different poses and when facing different directions. 


\section{Coupled Network with Gated Feature Extraction and Deformable Occlusion Handling}
\label{sec:proposedmethod}



Our motivation is to build a pedestrian detection framework which is robust for detecting pedestrians of small scale and with possible occlusions. In order to jointly achieve these two goals, the proposed detection network couples two sibling sub-networks, that is, a gated multi-layer feature extraction network and a deformable occlusion handling network.



Fig. \ref{fig:coupled_network} shows the block diagram of the proposed pedestrian detection framework. The VGG16 network \cite{VGG16} is employed as the {backbone network}. It contains 13 convolutional layers which can  be  regarded  as  five  convolutional blocks, i.e., $Conv1$, $Conv2$, $Conv3$, $Conv4$, and $Conv5$. The Region Proposal Network (RPN) is used to generate a number of pedestrian proposals which will be carefully examined. There are two coupled sub-networks which are built on top of backbone network and used for jointly learning and inference. 
The upper sub-network is the gated multi-layer feature fusion network which can adaptively provide discriminative features for each pedestrian candidate, and the lower sub-network is a deformable occlusion handling network which uses deformable RoI-pooling to produce robust features for occluded pedestrians.
In order to balance the contribution of the two sub-networks in training as well as inference stages, we perform normalization using an additional convolution layer at each sub-network and use element-wise summation for network coupling. The entire detection framework can be end-to-end trained.





In the following two sub-sections,  we will describe the implementation details of the gated multi-layer feature extraction sub-network and the regional occlusion handling sub-network, respectively. 

\begin{figure*}[tp]
    \centering 
    \includegraphics[width=1.9\columnwidth]{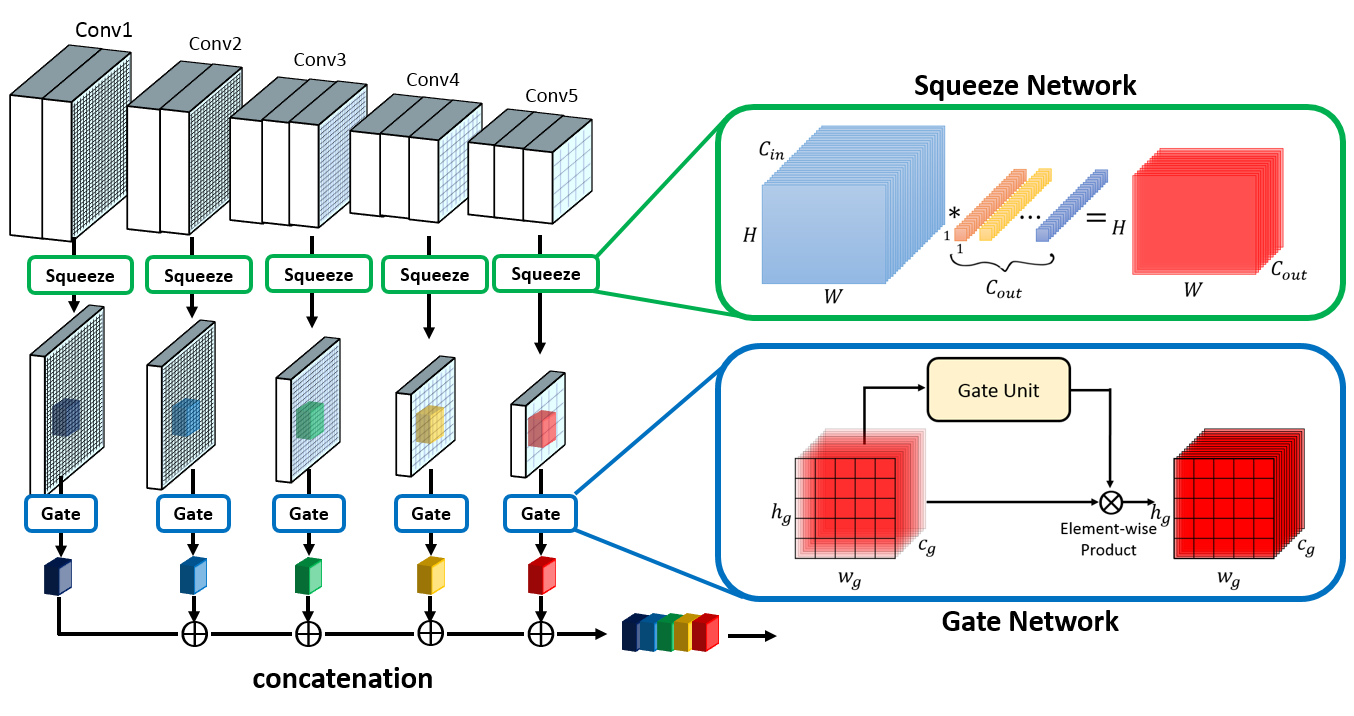}
    \caption{Overview of the proposed gated multi-layer CNN feature extraction network. Feature maps from each convolutional block of the backbone network are first compressed by the \textbf{squeeze networks} for dimension reduction. The squeezed lightweighted feature maps are then passed through \textbf{gate networks} for feature selection, and are integrated at the concatenation layer. The gate units will manipulate the CNN features to highlight the most suitable feature channels or feature components for a particular RoI, while suppressing the redundant or unimportant ones.
}
    \label{fig:gated_squeeze_multilayer_fusing} 
\end{figure*}

\subsection{Gated Multi-Layer Feature Extraction Sub-network}
\label{sec:gate_subnet}

The large and small size pedestrians in a captured image differ in two main aspects, that is, spatial resolution and visual appearance. Feature maps from different layers of a Convolutional Neural Network (CNN) have different reception field and spatial resolution and represent different levels of abstraction. Therefore, for pedestrians of different sizes, features that can best balance the representation ability and the feature resolution should come from different CNN layers.
For this reason, it is a good intuition to use different feature representations for detecting objects of different sizes. In \cite{Tianrui-BMVC2018}, the authors presented a Scale-Aware Multi-resolution (SAM) method which achieves a good feature representation by choosing the most suitable feature combinations for pedestrians of different scales. However, the multi-layer feature combinations are not automatically learned and there can only be a limited number of simple combinations of features to be tested.

In this paper, we aim at investigating a more advanced approach which can learn to select the best multi-layer feature combination for detecting pedestrians of various sizes. 
The proposed gated multi-layer feature extraction sub-network takes the features from all the five convolutional blocks as the input and will thereafter select the most discriminative feature components for different pedestrian candidates based on features from different layers. 
The gated multi-layer feature extraction sub-network realizes an automatic re-weighting of the multi-layer features from different layers of the backbone network using gate units.
Nevertheless, the gated network requires additional convolutional layers which induce a deeper RoI-wise sub-network at the cost of a higher complexity and a higher memory occupation. To remedy this issue, our gated sub-network includes squeeze units which are used to reduce the dimension of feature maps.

As illustrated in Fig. \ref{fig:gated_squeeze_multilayer_fusing}, features maps from each convolutional block of the backbone network are first compressed by a \textit{squeeze network}, then the RoI features pooled from the squeezed feature maps are passed through \textit{gate networks} for feature selection, and are integrated at the concatenation layer.

\subsubsection{Squeeze Network} 

A squeeze network is used to transform the input feature maps $\bm{F} = \left[{\bm{f}}_{1}, \cdots, {\bm{f}}_{C_{in}}\right] \in \mathbb{R}^{H \times W \times C_{in}}$ to lightweighted feature maps 
$\widehat{\bm{F}} = \left[\widehat{\bm{f}}_{1}, \cdots, \widehat{\bm{f}}_{C_{out}}\right] \in \mathbb{R}^{H \times W \times C_{out}}$ 
through $1 \times  1$ convolution:
\begin{equation}
    \widehat{\bm{f}}_i = \bm{v}_i * \bm{F},
\end{equation}
where $\bm{v}_i$ is the $i$-th learned filter in the squeeze network for $i=1,\cdots,C_{out}$, and `$*$' denotes convolution.






The reduction ratio of the squeeze network is defined as $r = \frac{C_{in}}{C_{out}}$.
We will show in Section \ref{sec:sqeeze_ratio} that by properly selecting the reduction ratio $r$ the squeeze network can reduce the RoI-wise sub-network parameters without noticeable performance downgrading.
Taking VGG16 $Conv4$ feature maps as an example, $Conv4$ feature maps which are of $C_{in}=512$ channels can be squeezed by a ratio of $r=2$ so as to be reduced to the lightweighted feature maps with $C_{out} = 256$ channels.


The RoI pooling will be performed on the squeezed lightweighted feature maps. The RoI pooled features will then be passed through a gate unit for feature selection. 


\subsubsection{Gate Network}

\label{sec:Gate_units}

A gate network will be used to manipulate the RoI pooled features to highlight the most suitable feature channels or feature components for a particular RoI, while suppressing the redundant or unimportant ones. 
As shown in the block diagram in Fig. \ref{fig:gated_squeeze_multilayer_fusing}, the output of a gate unit is used to perform an element-wise product with the RoI pooled features, deciding how the input feature would be manipulated. 


%

In general, a gate unit consists of a convolutional layer, two fully connected (fc) layers and a Sigmoid function at the end for output normalization. The convolutional layer and fc layers are associated with ReLU activation functions for non-linear mapping.

Given RoI pooled feature maps $\bm{R}$, the output of a gate unit $\bm{G} \in \mathbb{R}^{h_g \times w_g \times c_g}$ can be expressed as:
\begin{equation}
    \bm{G}=\sigma\left(\bm{W}_{2} \delta\left(\bm{W}_{1} \delta\left(\bm{C}_{1} * \bm{R}\right)\right)\right),
\end{equation}
where $\sigma(\cdot)$ denotes the Sigmoid function, $\delta(\cdot)$ denotes the ReLU activation function  \cite{ReLU2010}, and $\bm{\theta} = \{ \bm{C}_1, \bm{W}_1, \bm{W}_2 \}$ are the learnable parameters of the gate unit. 

The output of a gate unit $\bm{G}$ will be used to manipulate the regional feature maps $\bm{R}$ through an element-wise product:
\begin{equation}
    \bm{\widehat{R}}= \bm{G} \odot \bm{{R}},
\end{equation}
where $\odot$ denotes the element-wise product. 


The manipulated features $\bm{\widehat{R}}$ have the same size as its input RoI pooled feature maps $\bm{R}$, and will have enhanced information that is helpful for identifying the pedestrian within this RoI.

We have designed two types of gate units based on how the RoI pooled feature maps will be manipulated, namely, a spatio-wise selection gate unit and a channel-wise selection gate unit. The expectation is that the features manipulated by the output of the spatio-wise selection gate unit and the channel-wise selection gate unit will be able to have increased inter-dependencies among different channels and among different spatial locations, respectively. Therefore, the manipulated features will be more adaptive to the content within this RoI in terms of spatial variances and visual appearance variances, respectively.


\begin{figure}[tp]
    \centering
    \includegraphics[height=0.38\columnwidth]{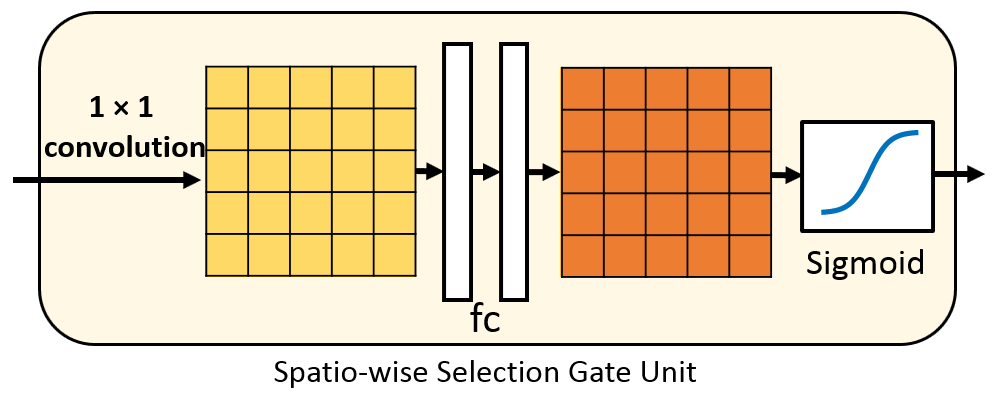}
    \caption{The gate unit for spatio-wise selection. In this gate unit, the (squeezed) RoI-pooled features will be transformed to a 2D map via $1 \times 1$ convolution. The 2D map is followed by two fully connected layers to generate another 2D map to be fed into the Sigmoid function. The output of this gate unit will be used to perform element-wise production in spatial domain with its input RoI features as given in Fig. \ref{fig:gated_squeeze_multilayer_fusing}.
    }
    \label{fig:gate_spatial}
\end{figure}


\textbf{Spatio-wise selection gate unit:} 
The output of a spatio-wise selection gate unit will be used to enhance the features at spatial locations where the features are essential for detecting pedestrians within this RoI. By reinforcing features at important spatial locations, the learned features are expected to be more robust for detecting pedestrians with partial occlusions.

The spatio-wise selection gate unit outputs a 2-dimensional (2D) map $\bm{G}$ of size $(h_g,w_g,c_g)=(h,w,1)$. It will be used to perform an element-wise product with the RoI pooled feature maps $\bm{R}$ which is of size $h \times w \times c$ through a $1 \times 1$ convolution. 

As shown in Fig. \ref{fig:gate_spatial}, through $1 \times 1$ convolution, the resulting 2D map has the same spatial resolution as the input feature. The 2D map is then passed through two fully connected (fc) layers and a Sigmoid function for normalization. The obtained 2D spatial mask $\bm{G}$ will be used to modulate the feature representation for every spatial location of the input feature. 
The feature values from all $C$ feature channels at spatial location $(i,j)$
will be modulated by the coefficient $\bm{G}(i,j,1)$.
	
	
\textbf{Channel-wise selection gate unit:}
The output of a channel-wise selection gate unit will be used to enhance the feature channels that are important for detecting pedestrians within this RoI. By reinforcing essential feature channels, the learned features are expected to be more adaptive to the feature resolution and visual appearance of the object.

\begin{figure}
    \centering
    \includegraphics[height=0.38\columnwidth]{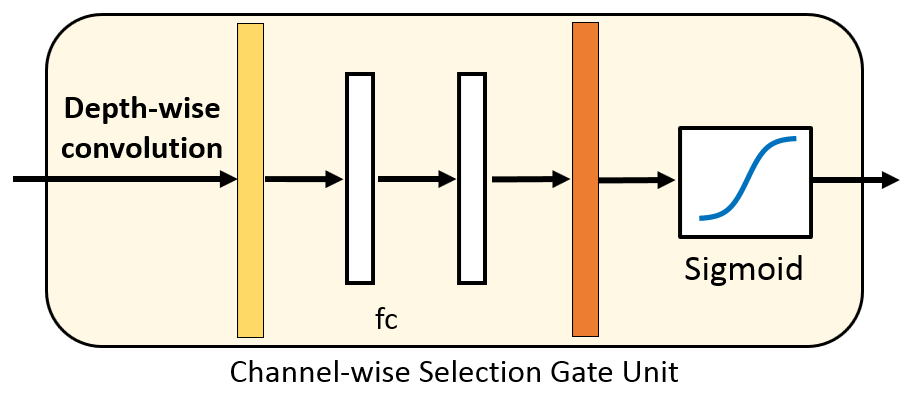}
    \caption{The gate unit for channel-wise selection. In this gate unit, the (squeezed) RoI-pooled features will be transformed to a 1D vector via depth-wise separable convolution. This vector is followed by two fully connected layers to generate another vector to be fed into the Sigmoid function. The output of this gate unit will be used to perform element-wise production in the channel direction with its input RoI features as in Fig. \ref{fig:gated_squeeze_multilayer_fusing}}
    \label{fig:gate_chn}
\end{figure}

The channel-wise selection gate unit generates a vector of size $(h_g,w_g,c_g) = (1,1,C)$ through $n$ {depth-wise separable convolution} \cite{MobileNetsV1}. As shown in Figure \ref{fig:gate_chn}, this vector is further passed through two fully connected layers and a Sigmoid function. The obtained $\bm{G}$ thereafter is used to perform a modulation with the convolutional features along the channel dimension. All the feature values within the $k$-th ($k\in [1,C]$) channel will be modulated by the $k$-th coefficient of $\bm{G}(1,1,k)$.

\begin{figure}[t]
    \centering \includegraphics[width=0.87\columnwidth]{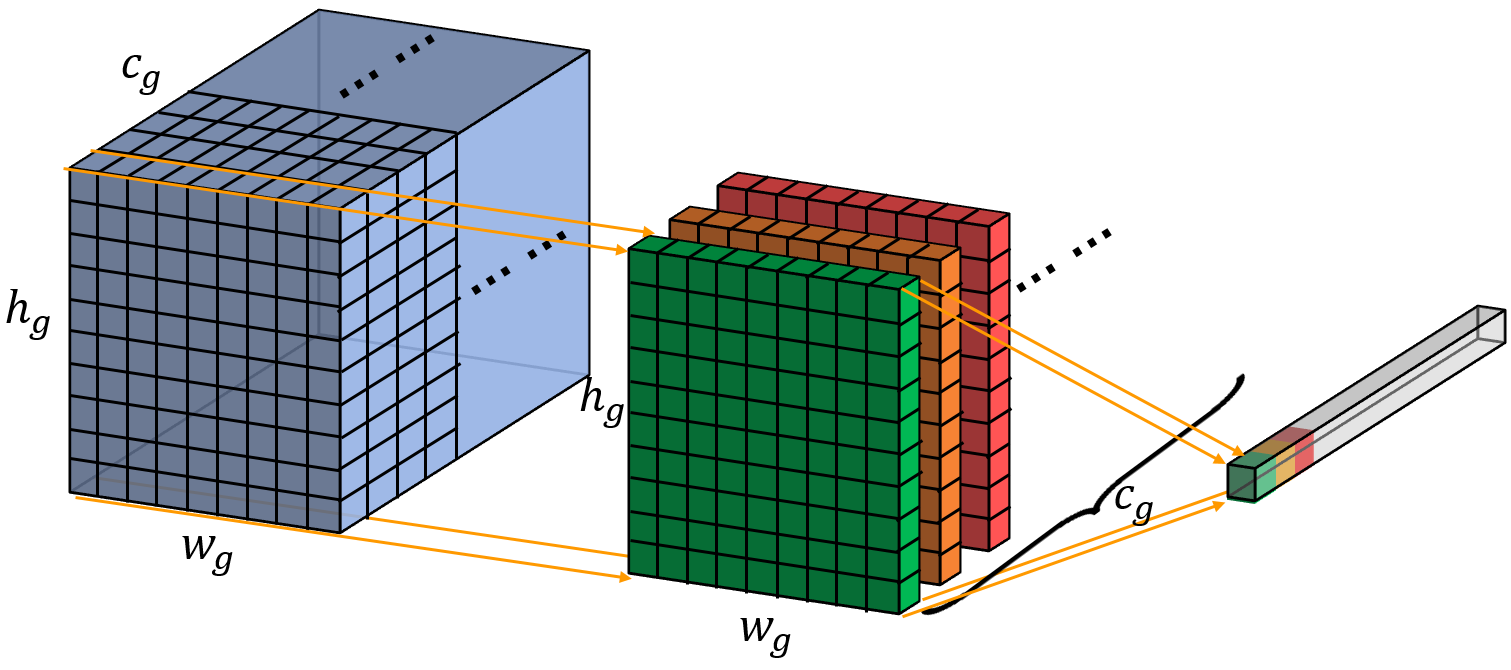}
    \caption{Illustration of depth-wise separable convolution. There are $c_g$ numbers of kernels applied separately onto the input feature map of size $h_g \times w_g \times c_g$. Each filter, of size $h_g \times w_g \times 1$, is convolved with a single channel of the input feature, resulting in a map of size $h_g \times w_g \times 1$. Then the separate output maps are stacked together to generate a vector of  $h_g \times w_g \times 1$.}
    \label{fig:depthwise_conv} 
\end{figure}

	
An illustration of depth-wise separable convolution is given in Fig. \ref{fig:depthwise_conv}. There are $c_g$ numbers of kernels applied separately on the input feature map of size $h_g \times w_g \times c_g$. Each filter, of size $h_g \times w_g \times 1$, is convolved with a single channel of the input feature, resulting in a map of size $h_g \times w_g \times 1$. Then the separate output maps are stacked together to generate a vector of  $h_g \times w_g \times 1$.


We will analyze and discuss in Section \ref{sec:experiments} that the proposed pedestrian detector will achieve different performance when used with different gate models.


\subsection{Deformable Occlusion Handling Sub-network}
\label{sec:occ_subnet}

To better solve the occlusion problem, we propose a sibling deformable occlusion handling sub-network with deformable regional RoI-pooling \cite{deformableConvnet2017} which will be able to generate robust features for pedestrians with partial occlusions. 
The deformable regional RoI-pooling performs regional RoI pooling on shiftable local grids which can better adapt to the position variations of the pedestrian parts, therefore will be capable of better handling occlusion problems in pedestrian detection.

\subsubsection{Baseline of the Occlusion Handling Sub-network}
\label{sec:RFCN}

The baseline network for this pipeline is the  Region-based Fully Convolutional Network (RFCN) \cite{RFCN}. The basic idea of the RFCN detector is based on the intuition that one can localize an object even only with partial information of the object.

\begin{figure}[t]
    \centering \includegraphics[width=0.9\columnwidth]{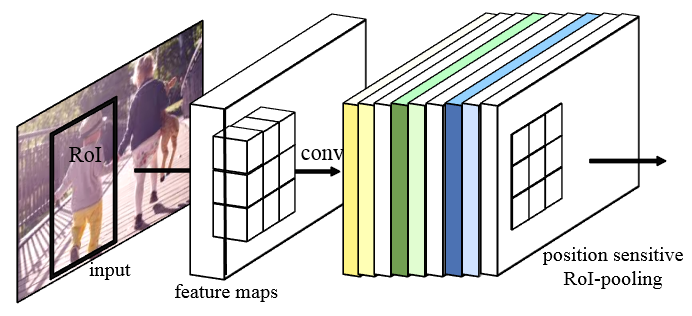}
    \caption{The block diagram of the RFCN detector \cite{RFCN} and an illustration of position sensitive feature maps. The RFCN detector divides a RoI into $k\times k$ local grids and generate $2k^2$ regional position sensitive feature maps for local parts. Here, we use $k=3$ for illustration purposes, and hence there are $3\times3=9$ spatial grids describing the relative position of each region with respect to the RoI, \textit{i.e.}, top-left, top-center, top -right, ..., bottom-right.}
    \label{fig:RFCN_featuremaps} 
\end{figure}


The RFCN detector considers different parts of an object separately by generating $C_{cls} \times k\times k$ regional position sensitive feature maps, each considering feature representation corresponding to an object class and at a grid location on a $k \times k$ grid. Fig. \ref{fig:RFCN_featuremaps} shows a block diagram of the RFCN detector in which we use $k=3$ for illustration, hence there are $3\times 3=9$ local grids in a RoI. The relative positions of each local grid with respect to the RoI are top-left, top-center, top -right, ..., bottom-right.

Let us denote with $\bm{S}_{(i,j)}^{c}$ the feature maps corresponding to class $c$ and grid location $(i,j)$.
The detection score is determined by combing the votes from these regional position sensitive feature maps. This operation is named Position-Sensitive RoI (PSRoI) pooling. 
PSRoI pooling enables the extraction of a fixed length feature representation for objects of arbitrary size from the $C_{cls} \times k^2$ regional position sensitive feature maps. For class $c$ and grid $(i,j)$, the extracted feature is obtained as the average of the feature maps $\bm{S}_{(i,j)}^{c}$ within grid $(i,j)$, i.e.,
\begin{equation}
    \bm{s}(c,i,j) = \frac{1}{n_{c,i,j}} \sum_{\bm{p} \in grid(i,j)} \bm{S}_{(i,j)}^{c}(\bm{p}_0 + \bm{p}),
\end{equation}
where $\bm{p}_0$ is set to the position of the upper-left grid, $\bm{p}$ enumerates spatial positions within the grid location $(i, j)$, and $n_{c,i,j}$ is the number of pixels in grid $(i,j)$ and is used as a normalization term.

\subsubsection{Occlusion Handling Sub-network Using Deformable Regional RoI-pooling}

\begin{figure}[t]
    \centering 
    \includegraphics[width=0.98\columnwidth]{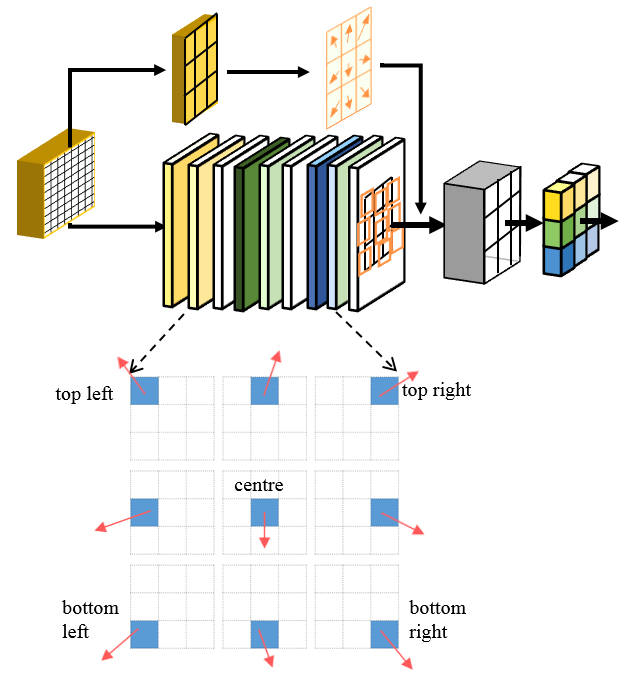}
    \caption{The proposed deformable occlusion handling network. Taking the CNN feature maps of an entire image as the starting point, a bank of $2k^2$ regional feature maps are generated by convolution. The following occlusion processing unit works with the regional feature maps to generate occlusion score map for each RoI.} 

\label{fig:deformable-occ-rfcn} 
\end{figure}


The proposed deformable occlusion handling sub-network applies deformable PSRoI pooling within regional-based fully convolutional network. Possible deformation and occlusion of the candidate pedestrian would be more adaptively handled through learned shiftable pooling grids for RoI pooling.
The schematic representation for our proposed occlusion handling sub-network is shown in Fig. \ref{fig:deformable-occ-rfcn}. 

Given the convolutional feature maps of the input image that have been extracted from the backbone network, as well as the candidate pedestrian proposal RoI output from the RPN network,  
the feature maps are first passed through a convolutional layer to generate a bank of regional position sensitive feature maps. For pedestrian detection, there are only $C_{cls}=2$ classes, i.e., pedestrian class and background class. Therefore, the deformable occlusion handling sub-network will generate $2k^2$ regional position sensitive feature maps, with $k^2$ feature maps responsible for the pedestrian class and the other $k^2$ feature maps for the background class.


We apply deformable PSRoI pooling \cite{deformableConvnet2017} to obtain a fixed length feature representation for each RoI. Deformable PSRoI pooling is more adaptive than the regular PSRoI pooling \cite{fastrcnn15} and has adaptive and shiftable pooling grids for each candidate pedestrian proposal.



The deformable PSRoI pooling uses shiftable pooling grids which can adapt to the local parts deformation of pedestrians. The extracted feature $\bm{s}(c,i,j)$ is the average of the feature maps $\bm{S}_{(i,j)}^{c}$ within a shifted grid: 
\begin{equation}
    \bm{s}(c,i,j) = \frac{1}{n_{c,i,j}} \sum_{\bm{p} \in grid(i,j)} \bm{S}_{(i,j)}^{c}(\bm{p}_0 + \bm{p}+\Delta p_{ij}),
\end{equation}
where offsets $\Delta p_{i j}$ for grid location $(i,j)$ of the RoI are learnt to shift the pooling grids.



As shown in Fig. \ref{fig:deformable-occ-rfcn}, the offsets $\Delta p_{ij}$ are learnt through additional convolutional layer (see the green pathway). The feature maps pooled from each shifted grid are used to vote for the detection score.

With the deformable PSRoI pooling, the regional features can be flexibly pooled from a set of RoI bins in order to better cover foreground instance (i.e. the candidate pedestrian) region. The localization capability can therefore be enhanced, especially for occluded and deformable pedestrians.





\subsection{Training the Coupled Network}

The gated multi-layer feature extraction sub-network and the deformable occlusion handling sub-network are coupled to achieve a simultaneous robust detection on small scaled pedestrians and partially occluded pedestrians. To have a balanced energy on the feature generated by two sub-networks, the feature representation from each sub-network will go through a $1 \times 1$ convolution before they are combined through element-wise summation. The combined feature representation will then be used for pedestrian classification and bounding box regression.

\subsubsection{Label Assignment} 
The training classification labels are assigned based on the overlapping ratio between a proposal and the ground-truth bounding box.
We assign a binary class label to each proposal for network training as in Faster-RCNN \cite{fasterRCNN2015}. A positive label will be assigned if it overlaps with any ground-truth proposal with Intersection-over-Union (IoU) higher than 0.7, or has the highest IoU with a ground-truth proposal. Proposals will be regarded as negative if the maximum IoU with all ground-truth proposals is lower than 0.3. 

\subsubsection{Loss Function} 
Pedestrian detection aims to determine whether the content within a candidate region is a pedestrian or not and estimate an accurate bounding box coordinate for pedestrian candidates.
The loss function contains a classification loss term and a regression loss term which are summed over all the $N$ proposal regions within a training batch data:
 \begin{equation}
    \mathcal{L} = \sum_{i=1}^N \mathcal{L}_{cls}(C^*_i,C_i) + \alpha\mathcal{L}_{reg}(B^*_i,B_i),
\end{equation}
where $C_i$ is the predicted probability of the $i$-th candidate bounding box region being a pedestrian and $B_i$ is the predicted bounding box coordinates; $C^*_i$ and $B^*_i$ are the ground-truth label of candidate region and the ground-truth bounding box positions, respectively; $\alpha$ is a scalar and is used to adjust the contributions of the two terms, and is empirically set as 1; $\mathcal{L}_{cls}(\cdot,\cdot)$ denotes the classification loss term  which is cross entropy loss over pedestrian class and non-pedestrian class, and $\mathcal{L}_{reg}(\cdot,\cdot)$ denotes regression loss term which is a smoothed $L_1$ loss, i.e.,
\begin{equation}
    \mathcal{L}_{\mathrm{reg}}\left(B^*,B\right)=\sum_{j \in\{\mathrm{x}, \mathrm{y}, \mathrm{w}, \mathrm{h}\}} \operatorname{smooth}_{L_{1}}\left(B_{j}^{*}-B_{j}\right),
\end{equation}
in which
\begin{equation}
    \operatorname{smooth}_{L_{1}}(x)=\left\{\begin{array}{ll}{0.5 x^{2}} & {\text { if }|x|<1}, \\ {|x|-0.5} & {\text { otherwise.}}\end{array}\right.
\end{equation}

\subsubsection{Optimization} 
The network weights of the backbone network (i.e., from $Conv1$ to $Conv5$ convolutional blocks) are initialized from the network pre-trained using the ImageNet dataset \cite{imagenet_dataset09}, while the network weights of other convolutional layers are initialized as a Gaussian distribution with mean 0 and standard deviation 0.01.
Stochastic Gradient Descent (SGD) with momentum is used to optimize the network weights of the propose pedestrian detection network. The learning rate of the algorithm is initialized at $1 \times 10^{-3}$ and was reduced by a factor of 10 for two times during the training. The momentum $\lambda$ is set to 0.9 and weight decay is set to $5\times 10^{-4}$. 
During training, a single image is processed in each mini-batch, and for each image there are 256 randomly sampled proposals used to compute the loss for this mini-batch. 
The whole network is fully convolutional and benefits from end-to-end approximate joint training and multi-task learning.

\section{Experiments}
\label{sec:experiments}

\subsection{Datasets and Evaluation Metrics}
\label{sec:settings}


\noindent \textbf{CityPersons:} CityPersons \cite{CityPersons} is a recent pedestrian detection dataset built on top of the CityScapes dataset\cite{Cityscapes} which is for semantic segmentation. The dataset includes 5, 000 images captured in several cities of Germany. There are about 35, 000 persons with additional $\scriptsize{\sim}$13,000 ignored regions in total. Both bounding box annotation of all persons and annotation of visible parts are provided. We conduct our experiments on CityPersons using the reasonable training/validation sets for training and testing, respectively. 

\begin{table}[]
\caption{The definition of reasonable, small, occlusion, and all subsets in CityPersons dataset.}
\centering
\begin{tabular}{c|c|c|c|c}
\hline
           & Reasonable & Small & Occlusion & All \\ \hline
Height     & $[50, \inf]$           & $[50, 75]$      &   $[50, \inf]$        &  $[20, \inf]$   \\ \hline
Visibility & $[0.65,1]$           & $[0.65,1]$      &  $[0.2,0.65]$          & $[0.2,1]$    \\ \hline
\end{tabular}
\label{tb:Subsets}
\end{table}

We evaluated the CityPersons dataset under four subsets of different ranges of pedestrian size and difference occlusion levels as shown in Table \ref{tb:Subsets}. Evaluations are measured using the log average miss rate (\textit{MR}) of false positive per image (FFPI) ranging from $10^{-2}$ to $10^{0}$ (denoted as \textit{MR}$_{-2}$).




\noindent \textbf{Caltech}: 
The Caltech-USA dataset \cite{PedBenchmark-2009CVPR-Dollar} and the improved annotations \cite{HowFar-2016} are used for training and evaluation. 
The training set contains 4,250 images which are obtained by extracting every the $30$-\textit{th} frame from the Caltech videos. 
The new high quality annotations provided by \cite{HowFar-2016} which correct some inaccurate annotations are used in our experiments.
The testing set contains $4024$ images of size $480\times640$. Following the evaluation of Caltech benchmark \cite{PedBenchmark-2009CVPR-Dollar}, only bounding boxes restricted in the range of $x\in[5, 635]$, $y\in [5, 475]$ are evaluated. We evaluate the detection performance on the following subsets: (1) Reasonable: height $[50,\inf]$, and visibility $[0.65, 1]$; (2) Medium: height $\in[30, 80]$ , and visibility $[0.2, 1]$; (3) Heavy: height $[50,\inf]$, and visibility $[0.2, 0.65]$.


\subsection{Experimental Setup}  

For the Citypersons dataset, we follow the same hyper-parameters in the Deform-ConvNet \cite{deformableConvnet2017} source code for fair comparison: the number of epochs is 7, the learning rate starts with $1\times10^{-3}$ and the scheduling step is at 5.333, the warm-up step is used with a smaller learning rate $1\times10^{-4}$ for 4,000 min-batches. Online Hard Example Mining (OHEM) is used for training. Among $N$ proposals, only the top $n$ (i.e., $n=300$) RoIs which have the highest loss are used for back-propagation.
For the Caltech dataset, we set the learning rate to $5\times10^{-3}$.

In order to reduce over-fitting we use data augmentation, which flips the images horizontally. All experiments are performed on a single TITAN X Pascal GPU.

\subsection{Effectiveness of Squeeze Ratio}
\label{sec:sqeeze_ratio}
The squeeze ratio $r$ affects the network in terms of feature capacity and computational cost. To investigate the effects of squeeze ratio, we conduct experiments using features from multiple convolutional layers that have been squeezed by $r=1,2,4,8, 16,32$. The performances are compared in Table \ref{tab:sq_ratio}.  

We find that squeeze network can reduce the RoI-wise sub-network parameters without noticeable performance deduction. We use the reduction ratio $r = 2$ as it is a good trade-off between performance and computational complexity. 

\begin{table}[t]
\caption{Missing rate (MR\%) on Citypersons validation set using different squeeze ratios $r$.}
\center \begin{tabular}{c|c|c|c|c}
\hline
squeeze ratio & Reasonable & Small & Occlusion & All   \\ \hline
$r=1$     & 14.49      & 39.65 & 56.97     & 43.7  \\ \hline
$r=2$      & 14.35      & 42.02 & 55.6      & 43.02 \\ \hline
$r=4$      & 14.63      & 44.33 & 56.34     & 42.93 \\ \hline
$r=8$     & 14.85      & 39.98 & 58.06     & 44.52 \\ \hline
$r=16$     & 14.72      & 40.18 & 53.97     & 43.07 \\ \hline
$r=32$     & 15.12      & 41.68 & 57.82     & 44.31 \\ \hline
\end{tabular} 
\label{tab:sq_ratio}

\end{table}

\subsection{Effectiveness of the Gated Multi-layer Feature Extraction Network}
\label{sec:gate_models_ablation_study}

We compare our gated multi-layer feature extraction network with a modified version of Faster-RCNN detector \cite{fasterRCNN2015} (we denote it as the \textit{Baseline1} detector thereafter in this paper). The Faster-RCNN detector was modified in terms of the region proposal network (RPN). For general object detection \cite{fasterRCNN2015}, a three-scale three-ratio anchor is used to generate 9 proposals at each sliding position. For pedestrian candidate proposal, we use anchors of a single ratio of $\gamma=0.41$ with $9$ scales. 

The \textit{Baseline1} detector \cite{fasterRCNN2015} only adopts the $Conv5\_3$ feature maps for feature representation and for the following classification and regression prediction. The limited feature resolution of $Conv5\_3$ restraints the capability for detecting small pedestrians. For our ``Spatio-wise gate" model and the ``Channel-wise gate" model in Table \ref{tab:gate_models_citypersons}, we use features extracted from the proposed gated multi-layer feature extraction network applying the two gate models respectively.  We dilate the $Conv5$ features by a factor of two for both the \textit{Baseline1} detector and our proposed gated sub-networks. The dilated convolution is proposed in \cite{SemanticDeepCFRs14} for better performance in semantic segmentation tasks. A dilated convolution ``inflates" the filter by inserting spaces between the kernel elements. The technique enlarges the receptive field without increasing the filter size and hence, induce little additional costs. Using dilate convolution 
is crucial for detecting small instances. 

As can be seen from Table \ref{tab:gate_models_citypersons}, both the spatio-wise gate model and the channel-wise gate model make improvements upon the \textit{Baseline1} detector. These results demonstrate the effectiveness of our proposed gated multi-layer feature extraction method. More specifically, the spatio-wise gate model makes more improvements on the ``\textit{Occlusion}" subset, while the channel-wise gate model makes more improvements on the ``\textit{Small}" subset.

For our experiments on the coupled network in Section \ref{sec:experiment_couple}, we select the channel-wise gate model to be used in the gated multi-layer feature extraction sub-network.
The reason is that the channel-wise gate model performs better on detecting small pedestrians, which is the main responsibility of the gated multi-layer feature extraction sub-network. 
The problem of detecting occluded pedestrians can be further addressed by the occlusion handling sub-network.

\subsection{Visualization of Gated Multi-layer Feature Selection}
\begin{figure*}[t]
\centering \includegraphics[width=1.95\columnwidth]{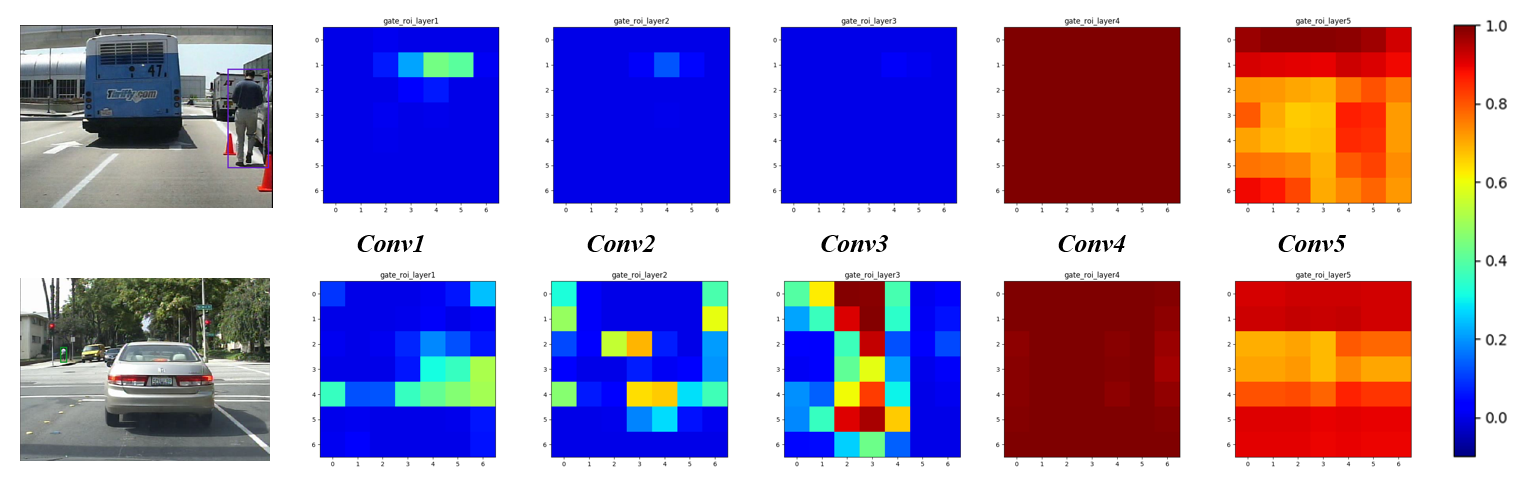}\caption{Examples of visualization of the gated multi-layer feature selection using spatial-wise model. The learned 2D gate masks for a small and a large pedestrian are different. For the small pedestrian, the gate model select a portion of features at the $conv1$ layer, which is  barely use by the large pedestrian. On the other hand, the large pedestrian selects to use more of features at the $conv5$ layer than the small pedestrian.}
\label{fig:2d-gate-mask} 
\end{figure*}

\begin{figure}[t]
\centering \includegraphics[width=0.98\columnwidth]{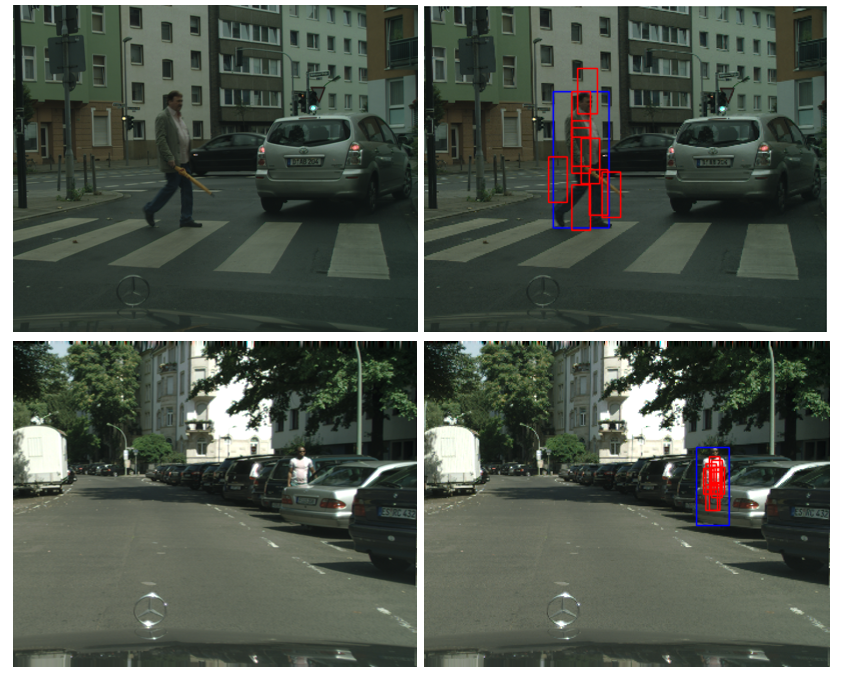}\caption{Examples of visualization of deformable RoI-pooling the occlusion handling sub-network for pedestrian detection. The upper example is a fully visible pedestrian, while the lower case is an occluded the pedestrian.}
\label{fig:deform-roi-eg} 
\end{figure}

An example is given in Fig. \ref{fig:2d-gate-mask} to visualize the spatio-wise gate selection. As described in section \ref{sec:Gate_units}, the output of the gate unit for spatio-wise selection is 2D masks. As can be seen from Fig. \ref{fig:2d-gate-mask}, the learned gate masks for a small and a large pedestrian are different. For a small pedestrian, the spatio-wise gate model selects a portion of features at the $Conv1$, $Conv2$ and $Conv3$ layer, which are barely used for the large pedestrian. On the other hand, the large pedestrian selects to use more of features at the $Conv4$ and $Conv5$ layer. The results, as has been expected, indicate that it is beneficial for small pedestrians to make use of feature representations from shallow layers which have higher resolution, while large pedestrian can benefit from features from deeper layers which have higher levels of abstraction.

\begin{table}[t]
\center
\caption{Comparison of pedestrian detection performance (in terms of MR\%) using different gate units on the Citypersons dataset.}
\begin{tabular}{c|c|c|c|c}
\hline 
Model  & Reasonable & Small & Occlusion & All   \\ \hline
FasterRCNN[\textit{Baseline1}] & 16.44  & 40.46  & 56.19  & 44.6 \\ \hline
Spatial-wise gate   & 13.64      & 41.17 & \textbf{52.37}     & \textbf{40.65} \\ \hline
Channel-wise gate & \textbf{13.49}   & \textbf{37.62}  & 53.53  & 41.76  \\ \hline
\end{tabular}

\label{tab:gate_models_citypersons}
\end{table}


\subsection{Effectiveness of Deformable RoI-pooling for Occlusion Handling}

We evaluate the effectiveness of  deformable RoI-pooling for occlusion handling by comparing its performance on the Citypersons dataset to our \textit{Baseline2} detector.

The Baseline2 detector is a modified version of the RFCN \cite{RFCN}. We re-implemented the RFCN detector \cite{RFCN} method using the VGG16 backbone network. Although the original RFCN detector in \cite{RFCN} utilizes the ResNet \cite{DeepResidualRecog2016} of 101 layers which achieves better results for general object detection than using the VGG16 network. We show in our experiments (see Table \ref{tab:RFCN_backbones}) that the performance for pedestrian detection using the ResNet is not as good as using VGG16. The reason is that the down-sampling rate for convolution layers in the ResNets is too large for the network to provide feature maps with sufficient resolution to detect small pedestrians. 
Moreover, we modified the RPN network for pedestrian candidate proposal in the same manner as we have done to the \textit{Baseline1} detector.


\begin{table}[]
\center
\caption{Comparisons of the performance (in terms of Missing rate MR\%, the lower the better) of the RFCN detector for pedestrian detection using different backbone networks, i.e, ResNet101 and VGG16 respectively.}
\begin{tabular}{l|l|l|l|l}
\hline
Models         & Reasonable & Small & Occlusion & All   \\ \hline
RFCN-VGG16     & 16.19      & 42.95 & 54.70      & 45.19 \\ \hline
RFCN-ResNet101 & 18.80       & 48.16 & 60.53     & 47.84 \\ \hline
\end{tabular}

\label{tab:RFCN_backbones}
\end{table}

The results of the RFCN \cite{RFCN} detector (in terms of miss rate) on the validation set with/without using deformable RoI-pooling layer are compared in Table \ref{tab:deformable-roi-pooling}. ``RFCN+deformable" denotes the improved version of RFCN which has been incorporated with deformable RoI pooling for occlusion handling.

As we can see, by applying deformable RoI-pooling, the improved RFCN detector has better performance in detecting pedestrian of the ``Small" and ``Occlusion" subsets. The overall performance, which can be evaluated using the ``Reasonable" and ``All" subsets, is therefore improved.

We give examples of visualization of deformable RoI-pooling method using in the occlusion handling sub-network for pedestrian detection in Figure \ref{fig:deform-roi-eg}. The upper example is a fully visible pedestrian, while the lower case is an occluded the pedestrian. 

\begin{table}[t]
\center
\caption{Comparisons of the performance of the RFCN detector for pedestrian detection with/without using deformable RoI pooling. ``RFCN+Deformable" denotes the improved version of RFCN which has been incorporated with deformable RoI pooling for occlusion handling.}
\begin{tabular}{l|l|l|l|l}
\hline
Methods         & Reasonable & Small & Occlusion & All   \\ \hline
RFCN-VGG16 [{Baseline2}]      & 16.19      & 42.95 & 54.70      & 45.19 \\ \hline
RFCN + Deformable   & 15.17      & 39.34 & 54.15     & 43.59 \\ \hline
\end{tabular}
\label{tab:deformable-roi-pooling}
\end{table}

\begin{table*}[tp]
\center
\begin{tabular}{l|l|l|l|l}
\hline
Methods     & Reasonable     & Small          & Occlusion      & All            \\ \hline
FRCNN+ATT+part \cite{occluded_shanshan2018} & 15.96 & - & 56.66 & - \\ \hline
CityPersons \cite{CityPersons} & 15.40 & - & - & - \\ \hline
TTL \cite{Small_ped_TTL2018} & 14.40 & - & 52.00 & - \\ \hline
CoupleNet [{Baseline3}]  & {14.36} & {38.56} & {52.95} & {42.25} \\ \hline
Repulsion Loss \cite{repulsion_loss2018} & 13.22 & 42.63 & 56.85 & 44.45 \\ \hline
OR-CNN \cite{Occ-awareRCNN2018} & 12.81 & 42.31 & 55.68 & 42.32 \\ \hline
CoupleNet+Gate+Occlusion \textbf{[Proposed]} & \textbf{12.37} & \textbf{\textbf{38.31}} & \textbf{49.81} & \textbf{40.39} \\ \hline
\end{tabular}
\label{tab:SoA-cityPersons}
\caption{Comparing the proposed detector to state-of-the-art methods (in terms of missing rate MR\%, the lower the better) on Citypersons validation set.}
\end{table*}

\subsection{Comparison to State-of-the-Art Pedestrian Detection Methods}
\label{sec:experiment_couple}

\subsubsection{CityPersons Dataset}

We compare the our proposed pedestrian detector with several state-of-the-art pedestrian detectors, including  \cite{occluded_shanshan2018},
FRCNN+ATT+part \cite{occluded_shanshan2018}, Adapted Faster RCNN \cite{CityPersons}, 
TTL \cite{Small_ped_TTL2018}, Repulsion Loss \cite{repulsion_loss2018}, OR-RCNN \cite{Occ-awareRCNN2018}.
The TTL \cite{Small_ped_TTL2018} method, is proposed for small-scale pedestrian detection using temporal feature aggregation. Repulsion Loss \cite{repulsion_loss2018} and OR-RCNN \cite{Occ-awareRCNN2018} are proposed to address the occlusion problems for pedestrian detection.
For fare comparison, all the performances are evaluated on the original image size ($1024\times 2048 $) of the CityPersons validation dataset.

We also implement a \textit{Baseline3} detector which is the coupled network of our Baseline1 detector and Baseline2 detector. That is, one of the sub-networks adopts the PSRoI pooling in Baseline2, while the other sub-network employs the RoI pooling as in the Baseline1 detector.

From Table \ref{tab:SoA-cityPersons}, we can observe that the proposed method surpasses the other approaches on all the subsets. The most notable improvements are on the ``Occlusion" and the ``Small" subsets. 
Our methods outperforms the state-of-the-art method OR-RCNN \cite{repulsion_loss2018} with an advantage of $6 \%$ on the ''Small" subset,  which highlights the effectiveness of our method for small-size pedestrian detection. In the case of the ``Occlusion" subset including some severely occluded pedestrians, our proposed method achieves the best $MR^{-2}$ performance of $49.81\%$), surpassing the second best pedestrian detector by a large margin of $5.87 \%$. These results clearly demonstrate the effectiveness of our proposed method for occlusion handling.

\begin{figure*}[tp]
    \centering 
    \includegraphics[width=2.05\columnwidth]{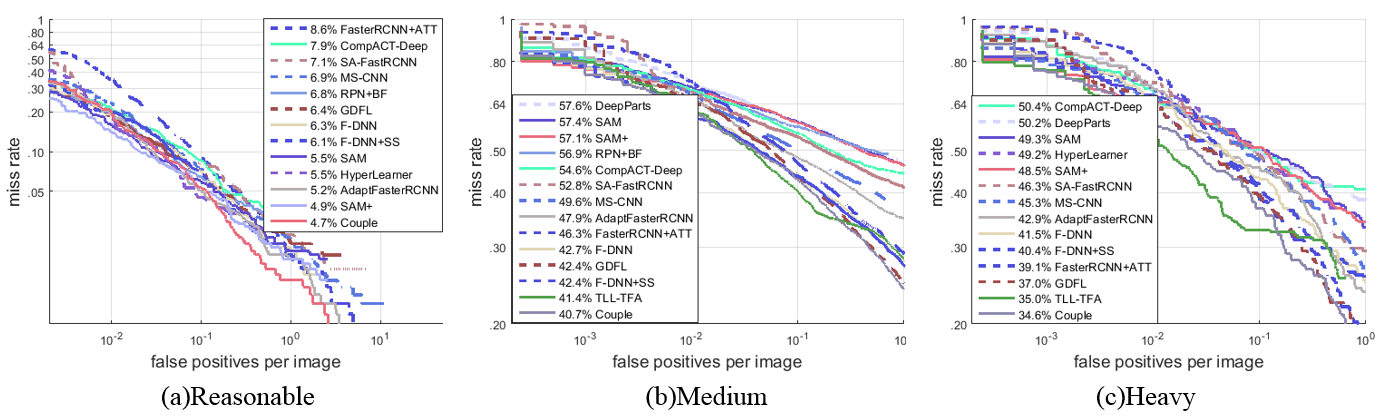}
    \caption{On the Caltech dataset under Reasonable, Medium and Heavy evaluation protocols, we compare our proposed method approach with some state-of-the-art methods.}
\label{fig:caltech_3graphs} 
\end{figure*}


\subsubsection{Caltech Dataset} 

In Figure \ref{fig:caltech_3graphs}, our pedestrian detectors are compared with the state-of-the-art pedestrian detection methods, namely, MRFC \cite{SemanticChannel}, CompACTDeep \cite{CompACTDeep}, SA-FastRCNN \cite{2018scale-awareRCNN}, MS-CNN \cite{MS-CNN2016} , RPN+BF \cite{IsfasterRCNNPed}, HyperLearner \cite{whatcanhelpPed}, OA-RCNN \cite{Occ-awareRCNN2018},  F-DNN/F-DNN+SS \cite{F-DNN}, FasterRCNN+ATT \cite{shanshan_occ_ped_cvpr2018},   GDFL\cite{GDFL_eccv2018} and TLL-TFA \cite{Small_ped_TTL2018}.
On the ``\textit{Reasonable}" subset, which only includes the pedestrian instances of at least 50 pixels in height and is widely used to evaluate the pedestrian detectors, the proposed method outperforms the second best method by 0.2 in terms of $MR_{-2}$. When it comes to the ``\textit{Small}" and ``\textit{Occlusion}" subsets, our method has achieved the best performance (i.e. $40.78\%$ and $34.60\%$), surpassing the previous state-of-the-art method GDFL\cite{GDFL_eccv2018} and PDOE \cite{PDOE_BiboxRegression_eccv2018} which are recent competitive methods on detecting small occluded pedestrians.
These improvements on the later two subsets indicate that our proposed coupled network is effective in detecting small-scale and occluded pedestrians.

\begin{figure*}[tp]
    \centering 
    \includegraphics[width=1.0\linewidth]{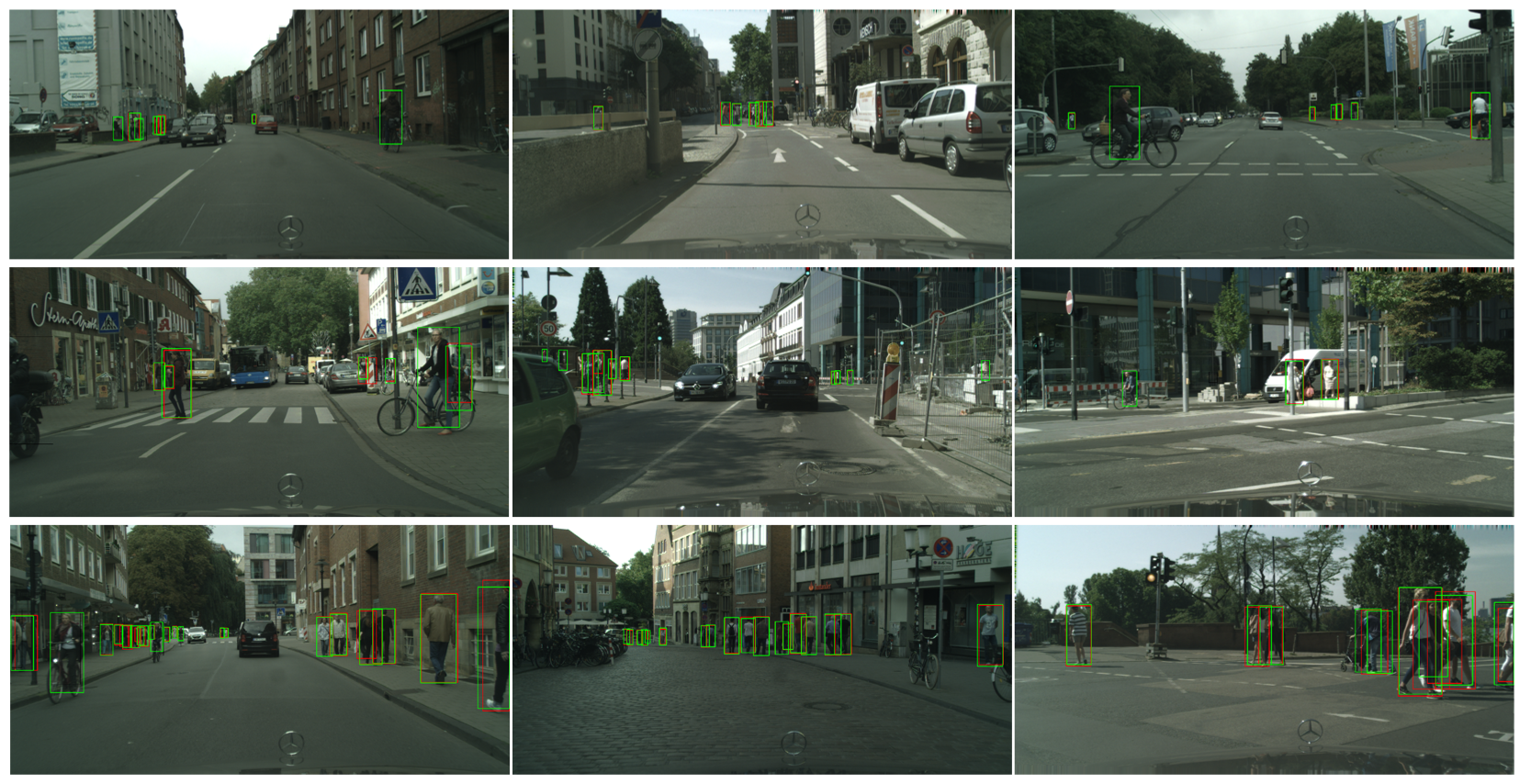}
    \caption{Qualitative results from our proposed detector on the CityPersons dataset. Green bounding boxes denote the ground-truth and red ones denote predicted bounding boxes. 
     The first and row give some detection examples of small-size pedestrians; the second and row give some detection examples of occluded pedestrians, and the third row shows some more challenging cases where pedestrians are crowed. The proposed pedestrian detection method performs well even on these challenging cases.
     }
\label{fig:qualitative_cityPersons} 
\end{figure*}


\subsection{Qualitative Results} 

\subsubsection{CityPersons Dataset} 
Fig. \ref{fig:qualitative_cityPersons} shows some exemplar detection results by our proposed pedestrian detector on the CityPersons dataset. In the first and second row, we show some examples of detected small pedestrians; in the third row we show some occlusion cases which have been successfully detected by our proposed method. From the qualitative results, we observe that our method is mostly successful on detecting even some small and heavily-occluded pedestrians. This demonstrates the effectiveness of our proposed coupled network on simultaneously addressing the problems of detecting small and occluded pedestrians in complex environment. 

\subsubsection{Caltech Dataset} 
\begin{figure*}[tp]
    \centering 
    \includegraphics[width=2.0\columnwidth]{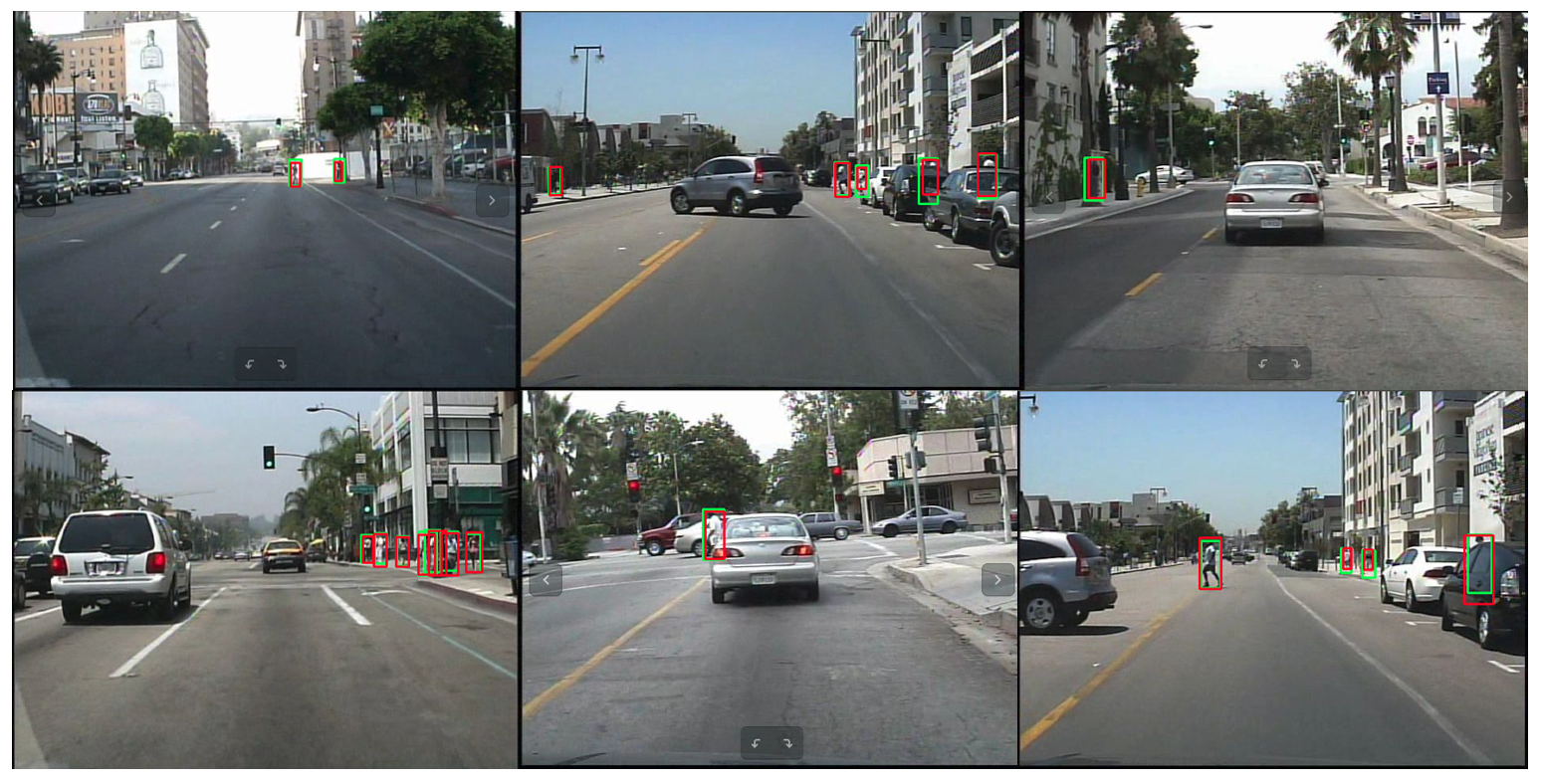}
    \caption{Qualitative results from our proposed detector on the Caltech dataset. Green and red bounding boxes denote the ground-truths and our predictions, respectively. We show some challenging cases where pedestrians are small and occluded.}
\label{fig:qualitative_caltech} 
\end{figure*}

We also show some qualitative results of our pedestrain detector on the Caltech dataset in Figure \ref{fig:qualitative_caltech}. Some challenging detection examples in this dataset are given. As can be seen, the proposed detection method can successfully detect small pedestrians even under crowed circumstances. For some extremely hard cases where the pedestrians are heavily occluded by cars and leaving merely the head region visualized, our pedestrian detector still performances well.   

\section{Conclusions}

In this paper, we proposed a robust pedestrian detection method with coupled network. The proposed coupled network consists of a gated feature extraction  sub-network which exploits different combinations of multi-resolution CNN features for pedestrian candidates of different scales and a occlusion handling sub-network which applies a deformable position sensitive pooling model. The two sub-networks provide two complimentary ways of detection to reinforce the robust detection results. Owing to the coupled framework, the proposed detection method can address the problem of detecting small and occluded pedestrians simultaneously. 
Extensive experiments on two widely used pedestrian detection datasets (i.e. CityPersons and Caltech) demonstrate the effectiveness of our proposed method for detecting pedestrians in urban scenarios, and show that it outperforms previous state-of-the-art methods by a large margin, where the largest improvement is registered for small-scale and heavily-occluded pedestrians. 

In the future, we can investigate the usage of temporal information since the pedestrian datasets are usually in video/ image sequences. The temporal information of consecutive sequences in the videos can provide extra contextual clues for better feature discrimination, especially for the occlusion cases where image-level information is relatively weak and ambiguous.  


\ifCLASSOPTIONcaptionsoff
  \newpage
\fi

\bibliographystyle{IEEEtran}
\bibliography{bibs}

\begin{thebibliography}{10}
\providecommand{\url}[1]{#1}
\csname url@samestyle\endcsname
\providecommand{\newblock}{\relax}
\providecommand{\bibinfo}[2]{#2}
\providecommand{\BIBentrySTDinterwordspacing}{\spaceskip=0pt\relax}
\providecommand{\BIBentryALTinterwordstretchfactor}{4}
\providecommand{\BIBentryALTinterwordspacing}{\spaceskip=\fontdimen2\font plus
\BIBentryALTinterwordstretchfactor\fontdimen3\font minus
  \fontdimen4\font\relax}
\providecommand{\BIBforeignlanguage}[2]{{%
\expandafter\ifx\csname l@#1\endcsname\relax
\typeout{** WARNING: IEEEtran.bst: No hyphenation pattern has been}%
\typeout{** loaded for the language `#1'. Using the pattern for}%
\typeout{** the default language instead.}%
\else
\language=\csname l@#1\endcsname
\fi
#2}}
\providecommand{\BIBdecl}{\relax}
\BIBdecl

\bibitem{IsfasterRCNNPed}
L.~Zhang, L.~Lin, X.~Liang, and K.~He, ``Is {F}aster {R-CNN} doing well for
  pedestrian detection?'' in \emph{European conference on computer
  vision}.\hskip 1em plus 0.5em minus 0.4em\relax Springer, 2016, pp. 443--457.

\bibitem{CityPersons}
S.~Zhang, R.~Benenson, and B.~Schiele, ``Citypersons: {A} diverse dataset for
  pedestrian detection,'' in \emph{Proceedings of the IEEE Conference on
  Computer Vision and Pattern Recognition}, 2017.

\bibitem{2018scale-awareRCNN}
J.~{Li}, X.~{Liang}, S.~{Shen}, T.~{Xu}, J.~{Feng}, and S.~{Yan}, ``Scale-aware
  {F}ast {R-CNN} for pedestrian detection,'' \emph{IEEE Transactions on
  Multimedia}, vol.~20, no.~4, pp. 985--996, April 2018.

\bibitem{Tianrui-BMVC2018}
T.~Liu, M.~Elmikaty, and T.~Stathaki, ``{SAM-RCNN}: Scale-aware
  multi-resolution multi-channel pedestrian detection,'' in \emph{British
  Machine Vision Conference (BMVC)}, September 2018.

\bibitem{Occ-awareRCNN2018}
S.~Zhang, L.~Wen, X.~Bian, Z.~Lei, and S.~Z. Li, ``Occlusion-aware {R-CNN}:
  detecting pedestrians in a crowd,'' in \emph{Proceedings of the European
  Conference on Computer Vision (ECCV)}, 2018, pp. 637--653.

\bibitem{DPM}
P.~F. Felzenszwalb, R.~B. Girshick, D.~McAllester, and D.~Ramanan, ``Object
  detection with discriminatively trained part-based models,'' \emph{Pattern
  Analysis and Machine Intelligence, IEEE Transactions on}, vol.~32, no.~9, pp.
  1627--1645, 2010.

\bibitem{pyramidSceneParsing2017}
H.~Zhao, J.~Shi, X.~Qi, X.~Wang, and J.~Jia, ``Pyramid scene parsing network,''
  in \emph{Proceedings of the IEEE conference on computer vision and pattern
  recognition}, 2017, pp. 2881--2890.

\bibitem{couplenet}
Y.~Zhu, C.~Zhao, J.~Wang, X.~Zhao, Y.~Wu, and H.~Lu, ``Couplenet: Coupling
  global structure with local parts for object detection,'' in
  \emph{Proceedings of the IEEE International Conference on Computer Vision},
  2017, pp. 4126--4134.

\bibitem{fasterRCNN2015}
S.~Ren, K.~He, R.~Girshick, and J.~Sun, ``Faster {R-CNN}: Towards real-time
  object detection with region proposal networks,'' in \emph{Advances in neural
  information processing systems}, 2015, pp. 91--99.

\bibitem{RFCN}
J.~Dai, Y.~Li, K.~He, and J.~Sun, ``R-{FCN}: Object detection via region-based
  fully convolutional networks,'' in \emph{Advances in neural information
  processing systems}, 2016, pp. 379--387.

\bibitem{Haar-ped1997}
M.~Oren, C.~Papageorgiou, P.~Sinha, E.~Osuna, and T.~Poggio, ``Pedestrian
  detection using wavelet templates,'' in \emph{cvpr}, vol.~97, 1997, pp.
  193--199.

\bibitem{SIFT}
D.~G. Lowe, ``Distinctive image features from scale-invariant keypoints,''
  \emph{International journal of computer vision}, vol.~60, no.~2, pp. 91--110,
  2004.

\bibitem{LBP-2002PAMI}
T.~Ojala, M.~Pietikainen, and T.~Maenpaa, ``Multiresolution gray-scale and
  rotation invariant texture classification with local binary patterns,''
  \emph{IEEE Transactions on Pattern Analysis and Machine Intelligence},
  vol.~24, no.~7, pp. 971--987, 2002.

\bibitem{DetectionandTracking-2007IJCV}
B.~Wu and R.~Nevatia, ``Detection and tracking of multiple, partially occluded
  humans by bayesian combination of edgelet based part detectors,''
  \emph{International Journal of Computer Vision}, vol.~75, no.~2, pp.
  247--266, 2007.

\bibitem{HOG}
N.~Dalal and B.~Triggs, ``Histograms of oriented gradients for human
  detection,'' in \emph{Computer Vision and Pattern Recognition (CVPR), 2005
  IEEE Conference on}, vol.~1.\hskip 1em plus 0.5em minus 0.4em\relax IEEE,
  2005, Conference Proceedings, pp. 886--893.

\bibitem{ACF_2014dollar2014}
P.~Doll{\'a}r, R.~Appel, S.~Belongie, and P.~Perona, ``Fast feature pyramids
  for object detection,'' \emph{IEEE Transactions on Pattern Analysis and
  Machine Intelligence}, vol.~36, no.~8, pp. 1532--1545, 2014.

\bibitem{FilterChannal-2015}
S.~Zhang, R.~Benenson, and B.~Schiele, ``Filtered channel features for
  pedestrian detection,'' in \emph{Computer Vision and Pattern Recognition
  (CVPR), 2015 IEEE Conference on}.\hskip 1em plus 0.5em minus 0.4em\relax
  IEEE, 2015, Conference Proceedings, pp. 1751--1760.

\bibitem{HowFar-2016}
S.~Zhang, R.~Benenson, M.~Omran, J.~Hosang, and B.~Schiele, ``How far are we
  from solving pedestrian detection?'' in \emph{Proceedings of the iEEE
  conference on computer vision and pattern recognition}, 2016, pp. 1259--1267.

\bibitem{2Ped-PAMI}
W.~Ouyang, X.~Zeng, and X.~Wang, ``Single-pedestrian detection aided by
  two-pedestrian detection,'' vol.~37, no.~9.\hskip 1em plus 0.5em minus
  0.4em\relax IEEE, 2015, pp. 1875--1889.

\bibitem{2PedTrack-IJCV-SiyuTang}
S.~Tang, M.~Andriluka, and B.~Schiele, ``Detection and tracking of occluded
  people,'' \emph{International Journal of Computer Vision}, vol. 110, no.~1,
  pp. 58--69, 2014.

\bibitem{MCF_TIP2017}
J.~Cao, Y.~Pang, and X.~Li, ``Learning multilayer channel features for
  pedestrian detection,'' \emph{IEEE transactions on image processing},
  vol.~26, no.~7, pp. 3210--3220, 2017.

\bibitem{Small_ped_TTL2018}
T.~Song, L.~Sun, D.~Xie, H.~Sun, and S.~Pu, ``Small-scale pedestrian detection
  based on topological line localization and temporal feature aggregation,'' in
  \emph{The European Conference on Computer Vision (ECCV)}, September 2018.

\bibitem{PDOE_BiboxRegression_eccv2018}
C.~Zhou and J.~Yuan, ``Bi-box regression for pedestrian detection and occlusion
  estimation,'' in \emph{Proceedings of the European Conference on Computer
  Vision (ECCV)}, 2018, pp. 135--151.

\bibitem{repulsion_loss2018}
X.~Wang, T.~Xiao, Y.~Jiang, S.~Shao, J.~Sun, and C.~Shen, ``Repulsion loss:
  Detecting pedestrians in a crowd,'' in \emph{Proceedings of the IEEE
  Conference on Computer Vision and Pattern Recognition}, 2018, pp. 7774--7783.

\bibitem{occluded_shanshan2018}
S.~Zhang, J.~Yang, and B.~Schiele, ``Occluded pedestrian detection through
  guided attention in cnns,'' in \emph{Proceedings of the IEEE Conference on
  Computer Vision and Pattern Recognition}, 2018, pp. 6995--7003.

\bibitem{shanshan_occ_ped_cvpr2018}
------, ``Occluded pedestrian detection through guided attention in cnns,'' in
  \emph{The IEEE Conference on Computer Vision and Pattern Recognition (CVPR)},
  June 2018.

\bibitem{GDFL_eccv2018}
C.~Lin, J.~Lu, G.~Wang, and J.~Zhou, ``Graininess-aware deep feature learning
  for pedestrian detection,'' in \emph{Proceedings of the European Conference
  on Computer Vision (ECCV)}, 2018, pp. 732--747.

\bibitem{RegionCNN-2014CVPR}
R.~Girshick, J.~Donahue, T.~Darrell, and J.~Malik, ``Rich feature hierarchies
  for accurate object detection and semantic segmentation,'' in \emph{Computer
  Vision and Pattern Recognition (CVPR), 2014 IEEE Conference on}.\hskip 1em
  plus 0.5em minus 0.4em\relax IEEE, 2014, Conference Proceedings, pp.
  580--587.

\bibitem{fastrcnn15}
R.~Girshick, ``Fast {R-CNN},'' in \emph{International Conference on Computer
  Vision ({ICCV})}, 2015.

\bibitem{PedBenchmark-2009CVPR-Dollar}
P.~Dollar, C.~Wojek, B.~Schiele, and P.~Perona, ``Pedestrian detection: A
  benchmark,'' in \emph{Computer Vision and Pattern Recognition (CVPR), 2009.
  IEEE Conference on}.\hskip 1em plus 0.5em minus 0.4em\relax IEEE, 2009,
  Conference Proceedings, pp. 304--311.

\bibitem{VGG16}
K.~Simonyan and A.~Zisserman, ``Very deep convolutional networks for
  large-scale image recognition,'' \emph{arXiv preprint arXiv:1409.1556}, 2014.

\bibitem{DPMareCNN}
R.~Girshick, F.~Iandola, T.~Darrell, and J.~Malik, ``Deformable part models are
  convolutional neural networks,'' in \emph{Proceedings of the IEEE conference
  on Computer Vision and Pattern Recognition}, 2015, pp. 437--446.

\bibitem{Multilabel_OCC_ped_ICCV2017}
C.~Zhou and J.~Yuan, ``Multi-label learning of part detectors for heavily
  occluded pedestrian detection,'' in \emph{The IEEE International Conference
  on Computer Vision (ICCV)}, Oct 2017.

\bibitem{ReLU2010}
V.~Nair and G.~E. Hinton, ``Rectified linear units improve restricted boltzmann
  machines,'' in \emph{Proceedings of the 27th international conference on
  machine learning (ICML-10)}, 2010, pp. 807--814.

\bibitem{MobileNetsV1}
A.~G. Howard, M.~Zhu, B.~Chen, D.~Kalenichenko, W.~Wang, T.~Weyand,
  M.~Andreetto, and H.~Adam, ``Mobilenets: Efficient convolutional neural
  networks for mobile vision applications,'' \emph{arXiv preprint
  arXiv:1704.04861}, 2017.

\bibitem{deformableConvnet2017}
J.~Dai, H.~Qi, Y.~Xiong, Y.~Li, G.~Zhang, H.~Hu, and Y.~Wei, ``Deformable
  convolutional networks,'' in \emph{Proceedings of the IEEE international
  conference on computer vision}, 2017, pp. 764--773.

\bibitem{imagenet_dataset09}
J.~Deng, W.~Dong, R.~Socher, L.-J. Li, K.~Li, and L.~Fei-Fei, ``{ImageNet: A
  Large-Scale Hierarchical Image Database},'' in \emph{Computer Vision and
  Pattern Recognition (CVPR), 2009 IEEE Conference on}, 2009.

\bibitem{Cityscapes}
M.~Cordts, M.~Omran, S.~Ramos, T.~Scharw{\"a}chter, M.~Enzweiler, R.~Benenson,
  U.~Franke, S.~Roth, and B.~Schiele, ``The cityscapes dataset,'' in \emph{CVPR
  Workshop on The Future of Datasets in Vision}, 2015.

\bibitem{SemanticDeepCFRs14}
L.-C. Chen, G.~Papandreou, I.~Kokkinos, K.~Murphy, and A.~L. Yuille, ``Semantic
  image segmentation with deep convolutional nets and fully connected crfs,''
  \emph{arXiv preprint arXiv:1412.7062}, 2014.

\bibitem{DeepResidualRecog2016}
K.~He, X.~Zhang, S.~Ren, and J.~Sun, ``Deep residual learning for image
  recognition,'' in \emph{Proceedings of the IEEE Conference on Computer Vision
  and Pattern Recognition}, 2016, pp. 770--778.

\bibitem{SemanticChannel}
A.~D. Costea and S.~Nedevschi, ``Semantic channels for fast pedestrian
  detection,'' in \emph{2016 IEEE Conference on Computer Vision and Pattern
  Recognition (CVPR)}, June 2016, pp. 2360--2368.

\bibitem{CompACTDeep}
Z.~Cai, M.~Saberian, and N.~Vasconcelos, ``Learning complexity-aware cascades
  for deep pedestrian detection,'' in \emph{The IEEE International Conference
  on Computer Vision (ICCV)}, December 2015.

\bibitem{MS-CNN2016}
Z.~Cai, Q.~Fan, R.~Feris, and N.~Vasconcelos, ``A unified multi-scale deep
  convolutional neural network for fast object detection,'' in \emph{ECCV},
  2016.

\bibitem{whatcanhelpPed}
J.~Mao, T.~X. andYuning Jiang, and Z.~Cao, ``What can help pedestrian
  detection?'' in \emph{Proceedings of the IEEE Conference on Computer Vision
  and Pattern Recognition (CVPR)}, 2017.

\bibitem{F-DNN}
X.~Du, M.~El-Khamy, J.~Lee, and L.~S. Davis, ``Fused dnn: A deep neural network
  fusion approach to fast and robust pedestrian detection,'' \emph{2017 IEEE
  Winter Conference on Applications of Computer Vision (WACV)}, pp. 953--961,
  2017.

\end{thebibliography}

\end{document}